\documentclass[runningheads]{llncs}
\usepackage[T1]{fontenc}
\usepackage{graphicx}
\usepackage{amssymb}
\usepackage{bbding}

\begin{document}
\title{Metanormative Theory for RL-Based Moral Agents}
%
%
\author{Aleks Knoks\inst{1}\Envelope\orcidID{0000-0001-8384-0328} \and
Marija Slavkovik\inst{2}\orcidID{0000-0003-2548-8623}}
\authorrunning{A. Knoks and M. Slavkovik}
%
\institute{University of Luxembourg, 2, place de l'Universit\'{e}, L-4365, Esch-sur-Alzette, Luxembourg \\
\email{aleks.knoks@uni.lu}\\ \and
University of Bergen, Fosswinckelsgt. 6, 5007 Bergen, Norway\\
\email{marija.slavkovik@uib.no}}
\maketitle              
\begin{abstract}
The overlapping disciplines of machine ethics and value alignment are concerned with designing artificial agents that are aligned with human values and that act in ethically acceptable ways. A recent trend in these disciplines is the use of reinforcement learning (RL) to design such agents, sidelining the philosophical literature that used to play a more central role. Against this backdrop, this paper pursues two goals. The first is to draw out ideas from recent work in metanormative theory that can be useful for designing artificial moral and value-aligned agents. The second is to examine the RL architecture through the lens of these ideas. This will give us clearer criteria for when an RL agent's behavior can be classified as moral, as well as a basis for evaluating and comparing different RL-based approaches to machine ethics and value alignment.

\end{abstract}
\section{Introduction}

In artificial intelligence (AI), an intelligent agent is classically understood as an entity that can perceive its environment and act upon that environment by taking actions toward achieving a goal \cite{Russell:2020}. In reinforcement learning (RL), intelligent agents take actions in a dynamic environment to maximize a reward signal \cite{Sutton2018}. After the training phase, an RL agent will typically have learned the strategy that maximizes this signal, and if all goes well, thereby also have learned how to achieve the user-specified goal. RL agents are thus designed to be rational in the narrow sense of being maximally efficient in achieving explicitly specified goals. But the more autonomy agents have and the wider their abilities, the more important it is for them to be capable of behavior and reasoning that is not only narrowly rational but also morally acceptable \cite{Chan23,conitzer2024automatemoraldecisionmaking}. In other words, it is at times important that AI agents, including RL agents, are also moral agents.  

Floridi and Sanders define an agent as a ``system [that is] situated within and [is also] a part of an environment, which initiates a transformation, produces an effect or exerts power on [the environment]'' \cite[p. 357]{floridi_morality_2004}. They further define \emph{moral} agents as those that are ``capable of morally qualifiable actions$\ldots$ or actions that can cause moral good or evil'' (p. 364). Given these definitions, RL agents are well-positioned to qualify as moral agents. 

One might be hesitant to accept Floridi and Sanders' definition---thinking, perhaps, that it overgeneralizes to even low-level animals and that agents whom we cannot hold responsible for their actions do not deserve the label `moral agents'. Nevertheless, one should recognize that even highly rudimentary agents can have both harmful and beneficial effects on people, and that we can meaningfully qualify (or assess) the actions of such agents as harmful and beneficial. And one should also recognize that it is in our best interests to ensure that rudimentary agents that can have harmful and beneficial effects on people are designed in such a way that their harmful effects are minimized and their beneficial effects are maximized. This is to say that Floridi and Sanders' definition of moral agents is best seen as stipulative. 

In any event, it invites two difficult engineering questions: (1) How are we to operationalize the notions of moral evil and moral good (or harmful and beneficial effects)? (2) And how are we to actually design agents that avoid moral evil and foster moral good? 

These questions are not new, as evidenced by the early work in machine ethics \cite{Anderson2007,Wallach2008}. What is new, however, is the developing research trend that attempts to answer both of these questions using the familiar tools of RL, typically overlooking work in moral philosophy and philosophical work more generally. The line of thought underlying this trend appears to run roughly as follows. Moral good and evil are nothing but further user-specified goals, and insofar as they are such goals, in an RL architecture they can be expressed using a separate reward function or combined with other user-specified goals into one overall reward function. Once this is done, standard RL techniques can be used to ensure that RL agents are moral. This is a sensible line of thought. However, it is also very general, and so it is no surprise that the research trend manifests itself as a wide array of very different proposals on how to use RL in the context of machine ethics, value alignment, and related areas. Furthermore, the criteria for classifying the behavior of RL agents as moral or value-aligned are not always clear.

Against this backdrop, this paper has two main goals. The first is to draw out some ideas from recent theoretical work in philosophy that can help us take important steps toward answering the above Questions (1) and (2). These ideas come from the field of philosophy called ``metanormative theory'' that is concerned with foundational questions about the nature of normativity and the structure of normative domains (including morality). The second goal is to examine the general architecture of an RL agent through the lens of these ideas. The hope is that this will give us clear criteria for when an RL agent's behavior can be classified as moral, as well as a sound basis for evaluating and comparing different RL-based approaches to machine ethics and value alignment.

Thus, we take the \textbf{main contribution} of this paper to consist in making some central insights from metanormative theory available to AI researchers working in machine ethics and value alignment, as well as illustrating how these insights can be used to evaluate approaches that rely on RL. The structure of the rest of this paper is as follows. Section \ref{sec:background} provides a quick review of the research areas that are most relevant to our purposes. Sections \ref{sec:normative-categories}--\ref{sec:central-concepts} present the ideas from metanormative theory that strike us as most relevant for designing artificial moral and value-aligned agents. Section \ref{sec:categories-and-RL} explains how these ideas apply to the standard RL architecture. Section \ref{sec:RL-examples} zooms in on three specific proposals for using RL in the design of value-aligned agents. Finally, Section \ref{sec:conclusion} concludes.

\section{Background}
\label{sec:background}

This section provides a snapshot of (1) machine ethics and value alignment; (2) metanormative theory; and (3) RL. It also briefly comments on the way these areas relate and distinguishes the perspective we take in this paper from the perspective of the normative multiagent systems (NorMAS) community.

\subsubsection{Machine Ethics and Value Alignment.} The discipline of \emph{machine ethics} is concerned with the behavior of machines toward people and other machines \cite{Anderson2007}. Since its inception, the field has been concerned with building artificial moral agents \cite{WallachCS2008}, and it seems fair to say that it has been dominated by logic-based methods \cite{Bello2023}.  
Machine ethics has conventionally drawn on ideas from moral philosophy, with its many attempts to operationalize various ethical theories, including utalitarianism, Kantian deontology, and virtue ethics---see, e.g. \cite[Part IV]{Anderson11}. \emph{Value alignment} is concerned with ensuring that the behavior of machines is aligned with human values \cite{gabriel_artificial_2020,ji_ai_2025}. Clearly, there is an overlap between the concerns of the two disciplines. Although their ostensible focuses and terminology are different, it remains to be seen whether there are any fundamental disagreements. Where disciplines may differ is their relation to moral philosophy, with machine ethics often focused on designing agents that align with particular moral theories, and value alignment less so \cite{Zhong25}. Although the value alignment literature does not see itself as concerned with the design of artificial moral agents, from the perspective of machine ethics, a machine that is aligned with human values would qualify as an artificial moral agent.

\subsubsection{Metanormative Theory.} The label `metanormative' generalizes the more familiar `metaethics' \cite{kaup24}. It is standard to distinguish first-order ethics from metaethics---see any good textbook, such as \cite{kaga98} or \cite{saher-land12}. The former is concerned with substantive accounts of \emph{moral} rightness, wrongness, goodness, badness, and the like, and is exemplified by such well-known ethical theories as utilitarianism, Kantian deontology, and virtue ethics. The latter explores the presuppositions of first-order ethical theories, as well as our moral discourse and practice \cite{sep-metaethics}. The work in metanormative theory is motivated by the observation that morality is not the only normative domain and the conviction that it is worthwhile to ask domain-neutral questions about basic normative concepts, categories, and their relationships \cite{kaup24}. Thus, metanormative theory can be characterized as the philosophical discipline that focuses on the presuppositions of normative theories (pertaining to moral philosophy, epistemology, practical rationality, and other areas of normative inquiry), as well as our everyday normative talk and practice.

While work in machine ethics and, to a lesser degree, value alignment draws on insights from first-order ethical theories, there has not been much overlap between these disciplines and metanormative theory. One noteworthy exception are approaches that appeal to the notion of (normative) reasons, see \cite{alca-etal24,cana-hort22,faroldi2026reasons}. As we are going to see, this notion plays crucial roles in metanormative theory. 

\subsubsection{Reinforcement Learning.} In an RL setting, we consider an agent interacting with a specified environment. The environment is typically modeled as a state transition system. A \emph{state} is represented by a set of variables (or propositions) whose values jointly determine the state of the environment. Depending on the setting, states may be fully or only partially observable to the agent. At each step, the agent selects one of a set of available \emph{actions}, each of which may, with a certain probability, cause a transition to a new state. In the standard formulation, the environment is modeled as a Markov decision process (MDP), meaning that the probability distribution over the next state depends only on the current state and the action chosen by the agent. The agent explores the environment with the aim of learning an \emph{optimal policy}, that is, a mapping from states to actions that maximizes the expected cumulative reward. The resulting policy is intended to produce the desired behavior as specified by the \emph{reward signal}. It is learned by optimizing this reward signal using a particular \emph{learning algorithm}. There are numerous learning algorithms that an RL agent can use to find an optimal policy.  What they have in common is that they guide the agent through trial-and-error interactions to learn the reward signal of each state-action transition. 

The reward signal can have a positive or a negative value---in the latter case it is a penalty. In a standard RL setting, the reward signal is assumed to be provided by the environment. Formally, it is the output of a \emph{reward function} that assigns reward values based on the current state, the action taken, and the resulting state. Reward functions can be learned or engineered \cite{Eschmann2021}. While in the standard RL, the agent learns the reward function, in Inverse Reinforcement Learning (IRL) \cite{Ng2000}, it starts out knowing the optimal policy and uses a learning algorithm to learn the reward function that would yield this policy.   

In sum, an RL scenario consists of a set of states, a set of actions available to the agent, a reward function (which may have an internal structure), and a learning algorithm that aims to identify an optimal policy.

The RL community is increasingly recognizing the need to train agents that accomplish their goals in an ethical and safe way -- see \cite{vishwanath2024}. But while its interest in ensuring that RL agents learn ``moral policies'' is increasing, the same cannot be said of its interest in moral (and other relevant areas of) philosophy. Thus, the standard approach to ensuring that the behavior of RL agents is ``moral'' is, in effect, to simply let humans train them or provide feedback on their behavior, without engaging with the questions of whether the training process is appropriate and whether the human-provided information is suitable. Conitzer et al. \cite{Conitzer2024} suggest that RLHF is growing to be the dominant approach here. However, they also admit that, despite its advantages, this approach faces some important challenges. We highlight two: the trained agent's moral behavior cannot be reproduced in new environments; and the agent is \emph{not} capable of any real justification of its decisions, beyond the uninformative ``it maximized the reward''. Here we note that, while we may readily accept ``I thought it was the right thing to do'' from a person who serves us vegan lunch, we would be less ready to take an artificial agent's word for it: what is good enough when it comes from people is often not when it comes from machines \cite{Hidalgo2021}.

Given this, we think it would be helpful to have clearer criteria for classifying an RL agent's behavior as moral. 
We also think that, given RL architecture, the moral status of an RL agent's behavior must ultimately depend on the moral status of some of the following: states; actions; reward signals; learning algorithms; and policies. To understand how to think about the moral status of these, we turn to metanormative theory. Before doing that, however, we distinguish our perspective from that of another related field.

\subsubsection{Norms and Normative Multi-Agent Systems.} Normative multi-agent systems (NorMAS) is the field that studies how rules, norms, and social expectations shape the behavior of multiple interacting agents. Specifically it is concerned with the representation and implementation of norms and the management of violations of norms in multi-agent system contexts, as well as the problem of enabling computational agents to reason with norms. While there is a connection between NorMAS, on the one hand, and machine ethics and value alignment, on the other, their focuses are different. For one, whereas NorMAS is concerned with the coordination among \emph{many agents} using norms, machine ethics and value alignment typically focus on an \emph{individual agent}. For another, there is also an important sense in which the scope of NorMAS is broader than that of machine ethics and value alignment, as the norms and rules it works with need not be moral or reflect human values.
There are further differences too.

NorMAS methods can be, and have been, used to further the goals of machine ethics and value alignment. Typically this means starting with a set of norms that are \emph{stipulated} to be moral, to be a kind of heuristic for moral values---here see \cite{montes2023}---or to reflect human values, and then studying agents that reason with these norms and try to comply with them. In this setup, it is very natural to call an agent's behavior ``moral'' insofar as it complies with the given set of norms and ``immoral'' insofar as it does not, and so we have clear criteria. It bears emphasis that, in this paper, we are \emph{not} presupposing that we are in a NorMAS setup, and that it is, at the very least, not obvious that the NorMAS setup reflects the perspective of metanormative theory on morality and other normative domains.

\section{Normative categories}
\label{sec:normative-categories}

The goal of this and the following section is to introduce the ideas from metanormative theory that, we think, should be in the toolbox of those working in machine ethics and value alignment. Although we present a fairly standard take on these ideas, it should be kept in mind that there is much disagreement among metanormative theorists, and that our presentation is deliberately simplified.

Consider an agent 
taking a decision in some morally
sensitive situation. For example, the agent might be deciding between buying a plane ticket or donating the money to Doctors Without Borders. There are many ways of conceptualizing the situation. However, we should agree on the following two minimal constraints that any sensible conceptualization must satisfy.
First, there has to be some \emph{standard} applying to the actions that the agent can perform or to the states of affairs that the agent can bring about in the situation. We get at this standard by using \emph{normative vocabulary}. For example, we could describe some action/state of affairs using the terms `right', `good', or `appropriate', and another using the terms `wrong', `bad', and `inappropriate'. Second, any sensible conceptualization of the situation has to distinguish between the features of the situation that are \emph{morally relevant} from the features that are not. For example, the fact that there are people dying in need of a doctor is, intuitively, morally relevant, while the facts that the doctors are on average 179cm tall and have brown eyes are not. Presumably, there is a tight relation between the morally relevant factors and  the standard that applies to actions/states of affairs, but what exactly this relation is is less clear. Ideally, the conceptualization of the situation would make it explicit.

As mentioned, we use a plethora of normative vocabulary to talk about morally sensitive situations---and normatively sensitive situations more generally. 
The literature in metanormative theory has done a lot to identify and classify the \emph{normative categories} this vocabulary picks out.\footnote{The question of what distinguishes normative from nonnormative categories is a difficult philosophical question. However, most of us have no trouble identifying normative categories using a kind of ``I know it when I see it'' test.}

\subsubsection{Right, Wrong, and Other Deontic Categories.} Two of the most important normative categories are \emph{right} and \emph{wrong}. For simplicity, we can think of a normatively sensitive situation as involving a set of exclusive and exhaustive actions available to the agent. Then the categories of right and wrong can be thought of as properties of the agent's actions. Also, notice that every action is either right or wrong, and that no action can be both right and wrong.\footnote{To be more precise, no action can be right and wrong, \emph{unless} it is \emph{right in some respect} and \emph{wrong in another}, or an action's rightness and wrongness are associated with \emph{different normative domains}. We discuss the qualifier ``in a respect'' and different normative domains in more detail in Section \ref{sec:central-concepts}.}

Right and wrong are the paradigm examples from the class of what the literature calls ``deontic categories''. This class also includes \emph{required}, \emph{obligatory}, \emph{forbidden}, \emph{prohibited}, \emph{permissible}, \emph{optional}, and their many cognates. Deontic categories are also often picked out using terms such as `ought to', `must', `may', and others. 
The deontic categories form a class because they resemble each other in several ways and because they are related to each other in ways that they are not related to nondeontic normative categories, cf. \cite{berker_22}.

\begin{table}
	\caption{Differences between the deontic and the evaluative categories.}\label{table1}
	\centering
	\begin{tabular}{c|c|c} 
		& Good/Bad & Right/Wrong \\ \hline\hline 
		Gradability & gradable & not gradable \\
		Neutral state & allow neutral states & no neutral states \\
		Privativity & nonprivative opposites & privative opposites \\
		Duality & not duals &  duals (of each other) \\
		Alternatives & depend on alternatives & do not depend on alternatives\\
	\end{tabular}
\end{table}

\subsubsection{Good, Bad, and Other Evaluative Categories.} In addition to the paradigm categories \emph{good} and \emph{bad}, the class of evaluative categories also includes \emph{neutral} (or \emph{indifferent}), \emph{better than}, \emph{worse than}, \emph{at least as good as}, \emph{no worse than}, \emph{weakly best}, \emph{strongly worst}, and many others. While some of these are naturally thought of as properties of actions available to the agent, others are relations between the available actions. Like the deontic categories, the evaluative categories form a class because they resemble each other in crucial ways. Berker \cite{berker_22} lists five features that distinguish evaluative categories from deontic ones. Table \ref{table1} illustrates these features by way of contrasting the categories \emph{good} and \emph{bad} with those of \emph{right} and  \emph{wrong}. Notice that `evaluative' is a cognate of `value(s)', and so that the evaluative categories can make the rather ambiguous notion of value---which, we submit, is overused in the ethics of AI and value AI literature---more precise.

\subsubsection{Justified, Unjustified, and Other Fittingness Categories.} The third class comprises fittingness categories, which include such categories as \emph{(in)appropriate}, \emph{(un)fitting},   \emph{(un)warranted}, \emph{(un)justified}, \emph{(im)proper},  and many others. Until recently, it was not common to separate fittingness categories as a distinct class from both deontic and evaluative categories. However, recent years have seen increased interest in fittingness, with some authors going as far as to argue that fittingness categories are more fundamental than the categories belonging to other classes--see, e.g. \cite{Howard2022-ROWFAU,howard2018fittingness,mchugh2016fittingness,mchu-way23}. We can think of fittingness categories as applying, in the first instance, to an agent's responses in normatively sensitive situations. McHugh and Way's \cite{mchu-way23} preferred phrase for describing fittingness categories is ``getting things right''. Note that a fitting response is meant to be different from a response that we would classify as right. Consider the example again: An agent who responds to the situation by donating the money to Doctors Without Borders \emph{because} people are dying gets things right, while an agent who responds to the situation by either buying a plane ticket or by donating the money \emph{because} most doctors have brown eyes, or \emph{because} it makes the agent look good, does not get things right. The fittingness categories share some features with the deontic categories and others with the evaluative ones, cf. \cite{berker_22}. 

\subsubsection{Reason-Related Categories.} The fourth class consists of categories that concern (normative) reasons. Examples include \emph{being a reason for}, \emph{being a reason against}, \emph{having decisive reason to}, \emph{being reasonable}, and a sequence of others. The categories \emph{being a reason for} and \emph{being a reason against} are the most important ones on the list. They are closely tied to what is called ``(dis)favoring relation'' \cite{scan98}. Thus, we can say that the fact that people are dying is a reason to donate or that it favors donating to Doctors Without Borders. It is almost universally agreed that the (dis)favoring relation is gradable, and that one reason can favor an action more than some other reason favors another action, cf. \cite{lord-magu16,Tucker2025}. 

If normatively sensitive situations ar thought of as involving a set of actions the agent can perform, it should be clear that the categories \emph{being a reason for} and \emph{being a reason against} are very different from deontic, evaluative, and fittingness categories: they are not properties of actions, they are not relations between actions, and they are not naturally thought of as properties of agents' responses to situations. Rather, it seems to be most natural to think of them as picking out some relata of the actions available to the agent, and it is tempting to identify reasons with the normatively relevant features of situations, or at least a subset of such features.
The categories of \emph{being a reason for}/\emph{against} play a central role metanormative theory. We illustrate this with two examples. First, it is widely accepted that they, along with some core deontic categories, are all we need to adequately capture the common structure of most (if not all) first-order ethical theories \cite{berk07,parf11,Tucker2025}. Once the common structure is identified, the substantive disagreements between these theories concern the features or facts that are identified as reasons. Second, there is the influential ``reasons first program'', the proponents of which hold that the notion of reason is basic and that other normative notions are to be analyzed in its terms \cite{danc93,parf11,raz90,scan98,schr21}.

Whiting's \cite{whit22} list of roles that reasons play in our normative theorizing and practices is also noteworthy. First, there is the \emph{explanatory role}: reasons explain such things as why a person ought to, may, or must act. Second, there is the \emph{justificatory role}: reasons justify (support, defend) actions, as well as serve as considerations that a person (or any agent) might cite in justifying their actions or another's actions. And finally, there is the \emph{guiding role}: one agent can point to a reason to guide another agent's action. Reasons can also guide action by figuring in a person's reasoning. (See \cite[p. 14ff]{whit22}.)

\subsubsection{Further categories.} 

The above four classes of categories have received the most attention in metanormative theory. But there are at least two others: the first is comprised of categories that have to do with virtue and vice; the second with categories relating to responsibility, blame- and praiseworthiness. They are sometimes called, respectively, the ``aretaic'' and ``hypological categories''. Although these categories are important, they seem to us to be of less direct relevance for the concerns of machine ethics and value alignment. Also, some authors hold that the latter class is a subset of that of fittingness---see e.g. \cite{berker_22}.

\section{Some Further Central Notions}
\label{sec:central-concepts}

Having laid out the most important normative categories, we discuss some further ideas and distinctions from metanormative theory which, we think, should be on the radar of those working in machine ethics and value alignment.

\subsubsection{Different Normative Domains.} In both everyday talk and research contexts, the actions agents take are sometimes described as ``moral'' and ``immoral'', or ``ethical'' and ``unethical''. Notice that these phrases are elliptical, saying that the actions in question are \emph{morally right}, \emph{morally good}, \emph{morally wrong} or some such. Notice also that it is easy to think of situations in which one and the same action taken by the agent is both \emph{legally} right and \emph{prudentially} right, but \emph{morally} wrong. This suggests that the use of `moral' and `immoral' signals that normative categories can pertain to different \emph{normative domains}---also called ``normative perspectives'' \cite{Tucker2025} or ``normative systems'' \cite{lord-magu16}---and this is the standard view in metanormative theory. It bears emphasis that the moral domain is often juxtaposed to the \emph{prudential domain} or the \emph{domain of self-interest}, which is concerned with an individual's own well-being, interests, and flourishing.

\subsubsection{Overall vs. Contributory.}  Metanormative theory standardly distinguishes between \emph{overall} and \emph{contributory} categories. A normative category is overall just in case it reflects all the normatively relevant features of the situation and the final assessment of the situation, and it is contributory just in case it contributes to this final assessment but does not by itself deliver it. 

The relation between the overall/contributory distinction and the four classes of normative categories is not perfectly straightforward. Reading some papers in metanormative theory leaves one with the impression that these distinctions completely cross-cut each other;
reading others, with the impression that the relation between them is more structured. 
We will not attempt to settle these differences.
For our purposes, it suffices that we acknowledge the significance of the distinction and recognize the paradigm examples of each. Thus, the core reason-based categories \emph{being a reason for} and \emph{being a reason against} are paradigm cases of contributory categories, and the deontic categories \emph{required}, \emph{permissible}, and \emph{right} are clearly overall categories, cf. \cite{lord-magu16}.

The overall/contributory distinction can be illustrated with the help of a staple example from \cite{sing72}: You can either keep your promise to meet your friend for dinner or save a drowning child from certain death. In this situation, it is overall right for you to save the child. It is (overall) required that you save the child and it is (overall) impermissible that you meet your friend for dinner instead. It is also (overall) good for you to save the child, and it is (overall) unjustified for you to meet the friend for dinner. Turning to the contributory categories now, the fact that the child is in mortal danger is clearly a reason for you to save her. Presumably, this reason also goes a long way toward explaining the presence of the above overall categories in this situation, that is, why it is impermissible and unjustified for you to meet the friend, letting the child die, and why it is (overall) good for you to save the child. However, notice that there is also a reason for you to meet your friend for dinner: the fact that you gave her a promise has not disappeared. While this reason gets outweighed by the other one, it can still figure in explanations of various sorts of thing we might want to say about the situation. For example, it can be used in an explanation of why you feel that you need to apologize to your friend even if you did what was right.

\subsubsection{The Qualifier ``In a Respect''.} Notice that many of the terms that we use to pick out both evaluative and deontic categories can be qualified with the phrase ``in a respect''. Taking this at face value might lead us to conclude that the qualified terms pick out contributory versions of the respective categories. With this, we could say that the fact that you will have kept your promise by meeting the friend in the above situation makes this action \emph{good in a respect} and that the fact that you will have let the child die makes this action \emph{bad in a respect}. Similarly, we could say that the action in question is  \emph{right in one respect} and \emph{wrong in another}. Most metanormative theorists readily accept qualified versions of evaluative categories, while only a small fraction of them accept qualified versions of the deontic ones---see \cite{whit22} and the response in \cite{kaup24}.

\subsubsection{The Qualifier ``All-Things-Considered''.} In the philosophical literature, normative terms are also sometimes qualified with the phrase ``all-things-considered''. This can mean one of two things. First, the phrase can be used synonymously with ``overall''. Second, it can signal---as it does in the context of the debate over whether there is an ``all-things-considered ought'' or ``ought simpliciter'' \cite{bake18,brow24}---that the term picks out a normative category that reflects a perspective of not a single normative domain, but, rather, the perspective of all (relevant) normative domains.  Thus, just as the overall moral categories are meant to reflect a perspective that takes into account all morally relevant features of the situation, the all-things-considered categories, in the sense in question, are meant to reflect a perspective that takes into account \emph{all} normatively relevant features of the situation, that is, morally relevant features, prudentially relevant features, as well as others. Whether such an all-encompassing normative perspective exists is controversial, as the differences in the positions of \cite{bake17} and \cite{brow24} illustrate.

\subsubsection{Normative Explanation and Justification.} 
Although the field of AI has not converged on how to define `explanation' and `justification', the following passage from a survey article reflects the way many AI researchers think about them: ``A key component of an artificially intelligent system is the ability to \emph{explain} the decisions, recommendations, predictions, or actions made by it and the process through which they are made. Explanation is closely related to the concept of \emph{interpretability}: systems are interpretable if their operations can be understood by a human, either through introspection or through a produced explanation. A related concept is \emph{justification}: intuitively, a justification explains why a decision is a good one, but may or may not do so by explaining how it was made'' \cite[p. 8]{biran2017explanation}.

Metanormative theory also often touches on issues pertaining to normative explanation and justification, although these notions are rarely central. V\"{a}rynen's recent work, \cite{vayr21,vayr-f}, is a welcome exception, and so we focus on it. 
Following \cite{vayr21}, we can think of normative explanations as explanations of why things (understood broadly to include facts and actions) have the normative features they do, or why they fall under certain normative categories. Thus, we can think of normative explanations as (veridical) answers to why-questions about normative categories in particular situations, that is, such questions as: Why is saving the child the best you can do? Why is it overall right for you to donate to Doctors Without Borders? And why is the fact that people are dying a reason to donate to Doctors Without Borders? 

Notice that the first two questions concern overall categories, while the third concerns a contributory one. Metanormative theorists agree that the general shape of answers to questions about the overall normative categories is fundamentally different from the general shape of answers to questions about the contributory ones. The general idea is roughly that the first kind of question needs to be answered by appeal to contributory normative categories, whereas answers to the latter kind of question need to appeal to substantive (normative) first-order theories. To illustrate, a successful answer to the question of why it is overall right for you to donate to Doctors Without Borders might point to the fact that people are dying. An adequate answer to the question of why the fact that people are dying is a reason to donate to Doctors Without Borders would need to appeal to something more substantial, such as bad consequences along utalitarian lines or failure to respect dignity along Kantian lines. It is important to notice here that the answer to the first question is perfectly compatible with either of the two fundamentally different answers to the latter question. This observation points to an idea that is widely agreed upon in metanormative theory: in many cases, the explanations of facts that concern overall normative categories are relatively independent of those that concern the contributory ones. So, when it comes to normative explanations, we see a kind of division of labor.

Turning to justification, we note that `justification' is a cognate of `justified', and that \emph{(un)justified} is a fittingness category. As a fittingness category, it has to do with responses and, more specifically, with responses that ``get things right''. So, justification is, roughly, what makes certain responses of the kind that get things right. V\"{a}yrynen's \cite{vayr21} thinking of justification as taking actions, attitudes, policies, norms, and such as their objects is well in line with this. 

Zooming out, we derive two conclusions from Sections \ref{sec:normative-categories} and \ref{sec:central-concepts}: 

\begin{itemize}
    \item Any use of moral vocabulary (e.g. `moral' and `ethical') to describe AI agents and their behavior must bottom out in terms of \emph{moral categories}, that is, normative categories pertaining to the moral domain. Similarly, any talk about values in AI should be made precise in terms of normative categories. 
    \item We should welcome those approaches to machine ethics and value alignment that can deliver explanations and justifications of facts that concern overall categories in terms of contributory ones (e.g. core reason-based categories).  
\end{itemize}

\section{RL Setup in Terms of Normative Categories}
\label{sec:categories-and-RL}

The main goal of this section is to explore how the normative categories from Section \ref{sec:normative-categories} apply to the general RL setup. We will look at some concrete proposals for using RL in the context of machine ethics and value alignment in Section \ref{sec:RL-examples}. 

It will be instructive to start by thinking of RL agents not against the backdrop of a moral standard (and moral categories), but against the backdrop of the standard of \emph{self-interest} or \emph{prudence}. In textbook presentations of RL, the user-specified goal is often presented as the RL agent's own, and the agent is presented as maximizing its own reward signal. For illustrative purposes, we take such descriptions at face value, as they give us a meaningful normative domain in an RL setup. 

Now we turn to the task of exploring how the normative categories apply in such a setup: a classical RL system with the reward function representing the agent's self-interest. Of all the elements of an RL system, the prime candidates for normative qualification are these three: (1) state-action pairs; (2) policies; (3) reward signals. 

Each state-action pair in an RL system is associated with a reward signal. This signal provides one straightforward way to make sense of state-action pairs in terms of core evaluative categories. Thus, a state-action pair is good (for the RL agent) if the corresponding reward signal is positive, bad (for the RL agent) if the signal is negative, and neutral otherwise. Clearly, we can also use the reward function to compare state-action pairs. So, it is meaningful to say that some of them are better, while others are worse. We can also contrast this local sense of `good', `bad', `better', and `worse' with a more global one that can be identified against the backdrop of optimal policy. Thus, there is a clear sense in which a locally bad state-action pair can nevertheless be good (for the agent) more globally, that is, if it is a part of the optimal policy. The optimal policy also makes for the most natural reference point for qualifying state-action pairs in terms of deontic categories, even though this might sound less natural here. Thus, we can say that a state-action pair is right (for the agent) and that it ought to be taken (given the agent's self-interest) insofar as it is part of the optimal policy. Similarly, we can make sense of state-action pairs in terms of fittingness categories by reference to the optimal policy: thus, it seems most natural to say that choosing some state-action pair is a fitting response on the part of the agent (given her self-interest) because it is a part of the optimal policy, and that choosing some (locally good) state-action pair is not a fitting response because it is not a part of the optimal policy.  Reason-related categories would seem not to apply to state-action pairs.

Similarly to state-action pairs, policies can be made sense of in terms of evaluative, deontic, and fittingness categories. Thus, we can say that some policies are good, bad, the best, the worst, right, wrong, justified, fitting, or unfitting for the agent (given her self-interests). Notice that we judge a policy as (un)fitting by reference to the reward function and the general aim of maximizing the signal: the optimal strategy is fitting (or justified) insofar and because it maximizes the signal associated with the reward function.

Next, we consider the reward signal. Given that it is gradable, it cannot be made sense of in terms of nongradable normative categories. This means that we should not be making sense of reward signals in terms of any of the deontic categories, nor any of the fittingness ones. This leaves us with two candidate classes: the evaluative and the reason-based. We can try to make sense of reward signals in terms of the core reason-based categories of being a reason for/against: Thus, we can say that the fact that the reward signal associated with some state-action pair $(s, a)$ is a negative real is a reason against $a$ in state $s$. Similarly, we can say that this fact is also a reason against any policy that includes $(s,a)$. We can also say that the reward signal associated with some other state-action pair $(s',a')$ is a positive real is a reason for $a'$ in $s'$. Similarly, we can say that this fact is also a reason for any policy that includes $(s',a')$.  

We register two problems with making sense of reward signals using reason-based categories: First, we should expect that, in many cases, there will be multiple reasons for and against the same action, not one. (This problem is less worrisome in a stochastic setting, where selecting an action in a state can lead to different outcomes and rewards. Also, it is not a problem when we think of reward signals as reasons for/against policies.) Second and more importantly, given that standardly both prudential and other normative reasons (promises, harms, etc.) are taken to come in \emph{different kinds}, we should expect reasons to be something more than mere numbers, which can always be added up. 

Alternatively, we can try to make sense of the reward signals in terms of evaluative categories. Above we saw that state-action pairs can be said to be good, bad, better, worse, etc. (for the agent). Accordingly, a positive reward signal can be thought of as the amount of (hitherto unspecified) goodness associated with a given state-action pair---and by extension, policy---and a negative reward signal can be thought of as the amount of (hitherto unspecified) badness associated with a given state-action pair---and by extension, policy. Presumably, the goodness here is some (still unspecified) positive value and the badness is some (still unspecified) negative value. If so, then of all the ideas relating to evaluative categories we discussed, reward signals most closely relate to \emph{goodness} and \emph{badness in a respect}. What is strange, however, is that there is only one respect in which actions can be good and only one respect in which they can be bad. Contrast this with the following case: you are choosing between smoking a cigarette or going for a run. Smoking is good for you insofar as it gives you immediate pleasure but bad for you insofar as it has a negative effect on your health. Going for a run, in turn, is bad for you insofar as it causes physical discomfort but also good for you insofar as it has a positive effect on your health. Clearly, the issue here runs parallel to what we saw when making sense of the rewards signal in terms of reason-based categories.

Notice that the reward function served as a basis for the (normative) domain of prudence. This function, together with the implicit call to maximize, gave us a standard, against the backdrop of which we could apply normative categories to various components of the RL system. It also pays to note how we would use the term `rational' in this context: we can talk about (more or less) rational actions and (more or less) rational behavior. However, all of these would seem to be derivative from what we might call the ``(most) rational policy'' (or ``policies''): the policy that is best, that is right, and that is fitting (given the RL agent's self-interests), or the policy that maximizes the reward signal.

Now we turn to the moral domain. From the perspective of metanormative theory, it is a separate domain that has a similar structure to that of self-interest, but is also independent: at times, what is best for an agent given her self-interests is morally bad---or vice versa---and what the agent rationally (or prudentially) ought to do is exactly what the agent morally ought not to do---or vice versa. How should one represent the moral domain in an RL system? The most natural and conservative answer runs thus: exactly the way we represented and made sense of the domain of self-interest. Accordingly, the first thing one needs is a distinct reward function that can serve as a basis for the moral domain.  
Against the backdrop of such a function, the normative categories apply to the components of an RL system in just the way they apply against the backdrop of a reward function representing self-interest. Thus, state-action pairs and policies can be made sense of in terms of deontic, evaluative, and fitting categories, and reward signals can be made sense of in terms of reason-based and evaluative categories. Notably, with respect to reward signals, we encounter the same problems: the setup allows for only one kind of moral reason or one kind of respect in which actions or policies can be morally good. The use of the term `moral' should run parallel to what we saw with `rational' above. We can elliptically talk about (more/less) moral actions and (more/less) moral behavior. However, these would be derivative from what we might call the ``(most) moral policy'' (or ``policies''): the policy that is morally best, morally right, and morally fitting, or the policy that maximizes the (moral) reward signal.

\begin{table}
	\caption{Applying normative categories to the components of RL.}\label{table2}
	\centering
	\begin{tabular}{l|c|c|c|c} 
		& Deontic & Evaluative & Fittingness & Reason-based \\ \hline\hline 
	State-action pairs, locally & $\checkmark$ & $\checkmark$ & $\checkmark$ & \\ 
		State-action pairs, globally & $\checkmark$ & $\checkmark$ & $\checkmark$ & \\ \hline 
		Policies & $\checkmark$ & $\checkmark$ & $\checkmark$ & \\ \hline
		Reward signals & & $\checkmark$ & & $\checkmark$ \\ 
	\end{tabular}
\end{table}

Before leaving this section, we summarize the way normative categories map onto the components of (basic) RL systems in Table \ref{table2}.

\section{Three Illustrative Examples}
\label{sec:RL-examples}

We contend that the theoretical framework developed above can be used to evaluate the promise of particular proposals for using RL to further the goals of machine ethics and value alignment. We do not have the space for a comprehensive analysis in this paper, but we can consider three illustrative examples.

\subsubsection{Abel et al.} In an early paper, Abel et al. \cite{Abel2016ReinforcementLA} ``argue that reinforcement learning achieves the appropriate generality required to theorize about an idealized ethical artificial agent, and [that it] offers the proper framework for grounding specific questions about ethical learning and decision making...''. They formalize ``ethical learning and decision-problems'' as solving \emph{partially observable} Markov decision processes (POMDPs) and advance their claims by analyzing two toy ``ethical dilemmas''. We focus on the first of these: ``Cake or Death''. The POMDP that Abel et al. work with involves a ``a single `true' ethical utility function'' which the agent can identify only through indirect observation (which is standard for POMDPs). Cake or Death is meant to model a situation in which the artificial agent is uncertain whether it is ``ethical'' to bake cakes for people or, rather, kill people. The POMPD contains three states: \emph{cake}, \emph{death}, and the terminal \emph{end}. In each of the nonterminal states, the agent can perform one of three actions: kill three people ($kill$), bake a cake for them $(bake\_cake)$, or ask a companion which action is ethical $(ask)$. Performing either of the first two actions leads to the terminal state, while asking leads to the same state, while revealing the rewards to the agent: $R(bake,bake\_cake) = 1$ and $R(death, kill) = 3$. Abel et al. consider three policies: immediately selecting $bake\_cake$; immediately selecting $kill$; and first asking what is ethical and then selecting $bake\_cake$ in $cake$ and $kill$ in $death$. They observe that, insofar as the third policy has the highest expected utility (given a uniform prior), it is also optimal and take this observation to suffice for the conclusion that the RL agent can identify the ``sensible'' policy.

It is not clear how Abel et al. think of the state where killing the three people leads to high reward: they might think of it as somehow representing a merely epistemic possibility or as part of a deliberately comic example. Either way, given our analysis from Section \ref{sec:categories-and-RL}, what they say implies that killing the three people is the morally best/right thing to do. By extension, killing the three people (in that state) is also part of any policy that is morally best, right, and fitting. 

Given our analysis, a notable strength of the approach of Abel et al. is that it works with a reward function (their ``true ethical utility function'') that serves as the cornerstone of the moral domain, and that makes it meaningful to apply moral categories in this context and classify a behavior as moral. On the negative side, the ``sensible'' policy that the RL agent ends learns is only one of the morally optimal policies. So, insofar as the goal was to show that RL can be used to reliably identify moral policies and induce moral behavior, it has not been met. Furthermore, the fact that the policy that the RL agent learns is moral appears to be incidental, as it is ``sensible'' insofar as it maximizes \emph{expected} reward signal. The phrase ``sensible moral policy'' might naturally suggest that the policy in question leads to moral behavior. But given our analysis, the behavior is moral if its underlying policy is morally good, right, or fitting, and the question of whether a policy is morally good, right, or fitting has little to do with maximizing \emph{expected} reward signal (as opposed to maximizing moral reward signal). The final and less significant point of criticism is that the moral theory that Abel et al. use in the example is very odd. Its odd character can be brought out by constructing an analogous case concerning self-interest: In it, the agent is uncertain between whether it is rational to eat nutritious food or commit suicide, and it learns the ``sensible'' policy of first asking what is rational and then eating nutritious food in one state and committing suicide in the other. The example implies that committing suicide is the rationally best/right thing to do, and that it is part of any policy that is best, right, and fitting, given the agent's self-interests. 

\subsubsection{Noothigattu et al.} Most RL-based approaches in value alignment envision an agent aiming to maximize some (nonmoral) reward under ethical (or other normative) constraints. One representative example is the work of Noothigattu et al. \cite{noothigattu-etal-2019} who try to teach AI agents ``ethical values'' with the help of RL and ``policy orchestration''. Their approach has three main components. The first is meant to help the agent learn constraints: first, IRL is used on demonstrations exhibiting desired behavior with the view of learning ``constrained rewards'', and then RL is used on these rewards to learn a ``strongly constrained'' (or ethical) policy $\pi_C$. The second component uses standard RL to learn a ``domain reward maximizing'' policy $\pi_R$. Finally, the third component is an algorithm that ``orchestrates'' the two policies, choosing one of them to play at each point in time. In \cite{noothigattu-etal-2019}, Noothigattu et al. illustrate the approach using the classic Pac-Man game. More specifically, they show that Pac-Man can do well in the game, while not eating any ghosts---presumably, the idea is that Pac-Man's behavior is ethical insofar as it does not eat ghosts. In the standard setup, that is, the setup in which Pac-Man learns the domain reward maximizing policy $\pi_R$, the rewards were as follows: $+10$ for eating a dot; $+500$ for eating all dots; $-500$ for colliding with a ``unscared'' ghost; and $+200$ for eating a ``scared'' ghost. And the (raw) rewards that were learned through IRL from the demonstrations and used for the constrained policy $\pi_C$, in turn, were as follows: $+.00284, +0.5507, -0.97059, -0.23434$. Noothigattu et al. note that the weight corresponding to eating ghosts (i.e. $-0.97059$) makes it clear that ``eating ghosts strongly violates the favorable constraints'' (p. 6380). Then they explain how the two policies $\pi_C$ and $\pi_R$ get mixed by the orchestrator, and how varying the hyperparamater controlling the trade-off between them results in Pac-Man either not eating ghosts or obtaining more points.

One might be tempted to call $\pi_C$ the ``ethical'' or ``moral policy'' (which Noothigattu et al. themselves do not do but seem to imply). However, given our analysis, we cannot talk about morally good/bad, right/wrong, etc. actions and morally good/bad, right/wrong, etc. policies here at all, because there is no moral reward function that could serve as the cornerstone for the moral domain. Furthermore, while the rewards that are used to obtain $\pi_R$ pertain to one domain---and hence, $\pi_R$ can be called the ``rational policy''---the rewards underlying $\pi_C$ blend domain rewards with (intended) moral penalties, creating a domain that is different from both that of morality and that of rationality. This means that there is no principled basis for the application of moral categories in this context. And from this, there is no basis for classifying behavior as moral.


Though the aim of constraining an RL agent that maximizes domain reward signal is clear, in the setup of Noothigattu et al. the boundary between morality and rationality gets blurred, and so it is not clear how to even apply moral categories here.

\subsubsection{Rodriguez-Soto et al.} In a series of papers \cite{rodriguez-soto_multi-objective_2023,rodriguezsoto2025morlincentives,Rodriguez-Soto2022Instilling}, culminating with \cite{rodriguezsoto2025morlincentives}, Rodriguez-Soto et al. use \emph{multi-objective} RL in the context of value alignment. They work with multi-objective Markov decision processes (MOMDPs) that include a \emph{vector} of reward functions where each function corresponds to a different ``value''. Notably, all of these values are meant to be ``ethical'', with one of them, the ``value of achievement'', representing the agent's ``individual objective''. The authors assume that these values form a ``value system'' or that they are totally ordered. They also assume that the value of achievement is never the top element in the value ordering. Since the values are totally ordered, they induce a lexicographic ordering on the rewards. So, an RL agent functioning in such an MOMDP learns to maximize the reward signal associated with the highest value, then with the second to highest value, and so on. Rodriguez-Soto et al. \cite{rodriguezsoto2025morlincentives} prove some important results about this setup and design some insightful  experiments, which are too complex to be discussed here.

This kind of setup is much better aligned with our analysis and metanormative theory: Policies that maximize a vector of reward functions representing ethical values are straightforward to make sense of in terms (deontic, evaluative, and fittingness) moral categories. Furthermore, the fact that there are now multiple reward functions goes some way toward resolving the issue with interpreting reward signals in terms of reasons/goodness in a respect: if there are multiple reward signals, we can talk about different moral reasons for actions/policies, or alternatively, different respects in which actions/policies can be good. What is unusual. That being said, the setup of Rodriguez-Soto et al. has at least two limitations. First, due to the fact that the reward functions, or ``values'', are totally ordered, the relative importance of reasons/respects of goodness is uniform. And this goes against the received wisdom in metanormative theory and other areas of philosophy. Here we may recall the drowning child scenario. While in it, the child's welfare was more important than keeping the promise, the received wisdom in philosophy is that, in other situations, the relative importance of these moral factors might be reversed. The second limitation concerns the value of achievement and its place in the value ordering. While Rodriguez-Soto et al. treat it as an ethical value, from the perspective of metanormative theory, it is much more natural to think of it as something that does not belong to the moral domain, but rather to the domain of self-interest. Thus, it is much more natural to think of it as a nonethical value. Now recall that the vector of reward functions corresponding to ethical values can be seen as giving rise to the moral domain. In light of this, the artificial agent would be naturally seen as selecting the most rational policy among the moral ones \emph{if} the value of achievement were always the least element in the value ordering. But as long as it is allowed to be higher in the ordering, the artificial agent is naturally seen to be designed to, at least sometimes, resolve trade-offs between more moral and more rational policies in favor of the more rational ones.

\begin{table}
	\caption{Strengths and Weaknesses of the Three Approaches.}\label{table3}
	\centering
	\begin{tabular}{p{1.75cm}|p{3.75cm}|p{6cm}} 
		Approach & Strengths & Main Weaknesses \\ \hline\hline 
		Abel et al. \cite{Abel2016ReinforcementLA} & The use of a ``true ethical utility function'' makes it meaningful to apply moral categories. & The agent learns only one of the morally optimal policies; the fact that it is moral is incidental; the agent's sole goal is to act morally. \\ \hline
		Noothigattu et al. \cite{noothigattu-etal-2019}  & The RL agent has a goal besides acting morally. & There is no principled basis for the application of moral categories and classifying a behavior as moral. \\ \hline 
		Rodriguez-Soto et al. \cite{rodriguezsoto2025morlincentives} & A vector of functions makes it meaningful to apply moral categories and makes room for different kinds of moral factors. & The relative importance of moral factors is context-invariant; the characterization of the function associated with  achievement as moral is dubious; the agent is sometimes trained to behave immorally.  \\
	\end{tabular}
\end{table}

Much more could be said about these three approaches---Table \ref{table3} summarizes their strengths and weaknesses---and there are many other RL-based approaches that deserve careful analysis. However, for reasons of space, such an analysis must be left for another day. We hope that this section has illustrated how the perspective of metanormative theory on RL can be used to evaluate and also compare competing RL-based approaches to machine ethics and value alignment.

\section{Conclusion}
\label{sec:conclusion}

This paper set out to contribute to meeting the challenge of designing agents that avoid moral evil and foster moral good. To this end, we provided a brief introduction to metanormative theory and its treatment of morality and other normative domains. We introduced the core normative categories, along with some other ideas that are central to this area. We also discussed the way these ideas apply to the RL architecture, as well as illustrated how the perspective of metanormative theory can be used as a criterion for evaluating and comparing approaches to machine ethics and value alignment. We are not suggesting that this criterion should be decisive. However, we do think that, ceteris paribus, an approach's promise increases with the extent to which it can be understood in terms of notions from metanormative theory.

%
%
%

\subsubsection{Acknowledgments.} We thank the participants of EMAS 2026 and the workshop \emph{Opportunities and Limitations of Artificial Moral Agents}, which took place in Utrecht in March 2026, for feedback on the material presented here. We also thank David Streit and the three anonymous referees for EMAS 2026 whose written comments have led to major improvements in this paper. Knoks was supported by the Luxembourg National Research Fund through the project \emph{The Epistemology of AI Systems} (C22/SC/17111440/EAI).

\subsubsection{Disclosure of Interests.} The authors have no competing interests.

 \bibliographystyle{splncs04}
 \bibliography{mybibliography}

\end{document}